\documentclass[10pt,twocolumn,letterpaper]{article}

\usepackage{cvpr}              

\usepackage{graphicx}
\usepackage{amsmath}
\usepackage{amssymb}
\usepackage{booktabs}

\usepackage{multirow}
\usepackage{multicol}
\usepackage{times}
\usepackage{epsfig}
\usepackage{amsfonts}
\usepackage{array}
\usepackage{caption}
\usepackage{subcaption}
\usepackage{arydshln}

\usepackage{balance}

\usepackage[pagebackref,breaklinks,colorlinks]{hyperref}

\usepackage[capitalize]{cleveref}
\crefname{section}{Sec.}{Secs.}
\Crefname{section}{Section}{Sections}
\Crefname{table}{Table}{Tables}
\crefname{table}{Tab.}{Tabs.}

\def\cvprPaperID{5559} 
\def\confName{CVPR}
\def\confYear{2022}

\begin{document}


\title{DALG: \textit{D}eep \textit{A}ttentive \textit{L}ocal and \textit{G}lobal Modeling for Image Retrieval}

\author{Yuxin Song$^{*}$, Ruolin Zhu$^{*}$, Min Yang$^{*}$, Dongliang He$^{*}$\\
{\tt\small songyuxinbb@outlook.com}
}

\maketitle

\begin{abstract}

   Deeply learned representations have achieved superior image retrieval performance in a retrieve-then-rerank manner. Recent state-of-the-art single stage model, which heuristically fuses local and global features, achieves promising trade-off between efficiency and effectiveness. However, we notice that efficiency of existing solutions is still restricted because of their multi-scale inference paradigm.  
   In this paper, we follow the single stage art and obtain further complexity-effectiveness balance by successfully getting rid of multi-scale testing. To achieve this goal, we abandon the widely-used convolution network giving its limitation in exploring diverse visual patterns, and resort to fully attention based framework for robust representation learning motivated by the success of Transformer. Besides applying Transformer for global feature extraction, we devise a local branch composed of window-based multi-head attention and spatial attention to fully exploit local image patterns. Furthermore, we propose to combine the hierarchical local and global features via a cross-attention module, instead of using heuristically fusion as previous art does. With our Deep Attentive Local and Global modeling framework (DALG),  
   extensive experimental results show that efficiency can be significantly improved while maintaining competitive results with the state of the arts. 
\end{abstract}

\section{Introduction}
\label{sec:intro}

Image retrieval aims at searching for similar images from an image gallery according to an image query. In this task, visual features (or descriptors) of images play the central role for similarity matching among query-candidate pairs. Many handcrafted features, such as SIFT \cite{lowe2004distinctive} and SURF \cite{bay2008speeded}, have been designed for image matching. These features, however, cannot well represent the global image information thus their retrieval performances are limited. 
Today, deep learning has dominated the trends of many computer vision tasks, so is the case for image retrieval \cite{babenko2014neural, arandjelovic2016netvlad, gordo2017end,revaud2019learning,noh2017large,ng2020solar,cao2020unifying,Yang_2021_ICCV}.  

\begin{figure}
    \centering
     \begin{subfigure}[b]{\columnwidth}
         \centering
         \includegraphics[width=\columnwidth]{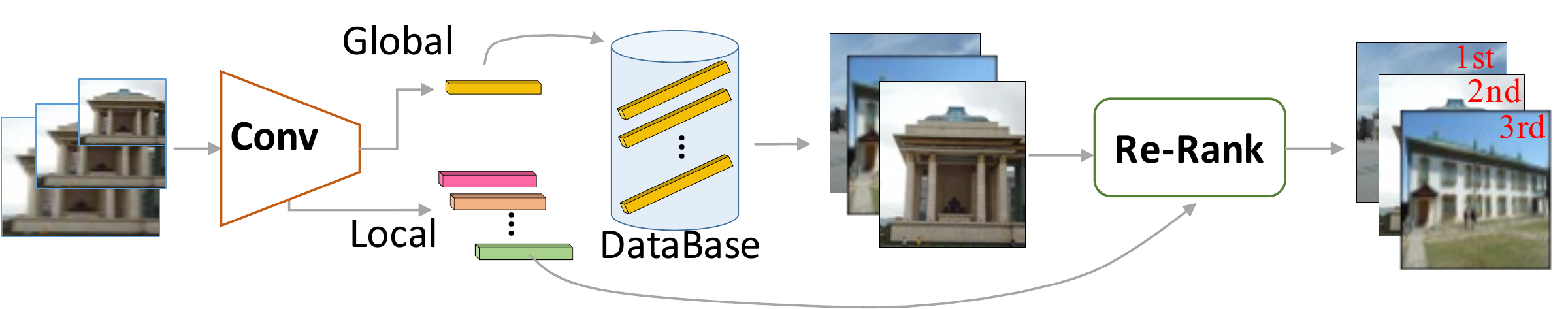} 
         \caption{Multi-scale testing pipeline of state-of-the-art two-stage solution DELG\cite{cao2020unifying}}
         \label{fig:delg}
     \end{subfigure}
     \begin{subfigure}[b]{\columnwidth}
         \centering
         \includegraphics[width=\columnwidth]{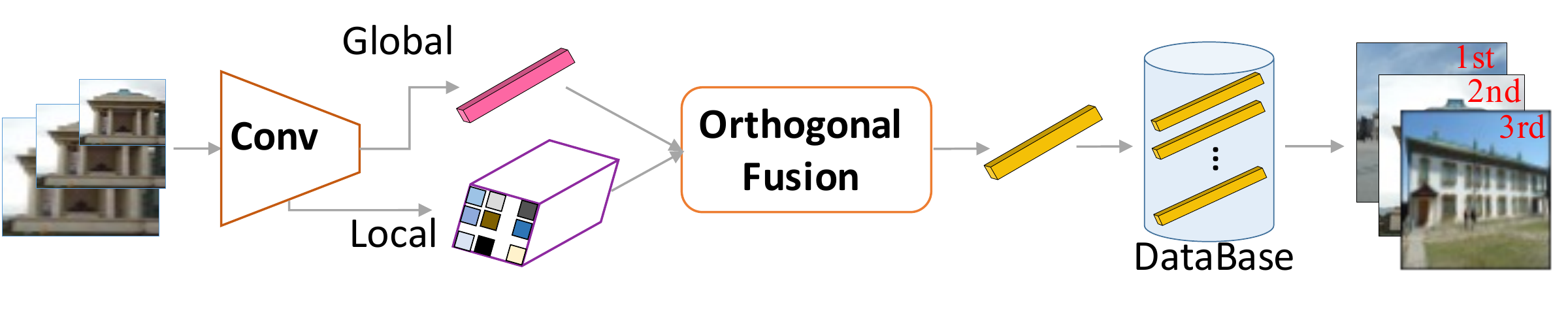} 
         \caption{Multi-Scale testing pipeline of state-of-the-art end-to-end solution DOLG\cite{Yang_2021_ICCV}}
         \label{fig:dolg}
     \end{subfigure}
     \begin{subfigure}[b]{\columnwidth}
         \centering
         \includegraphics[width=\columnwidth]{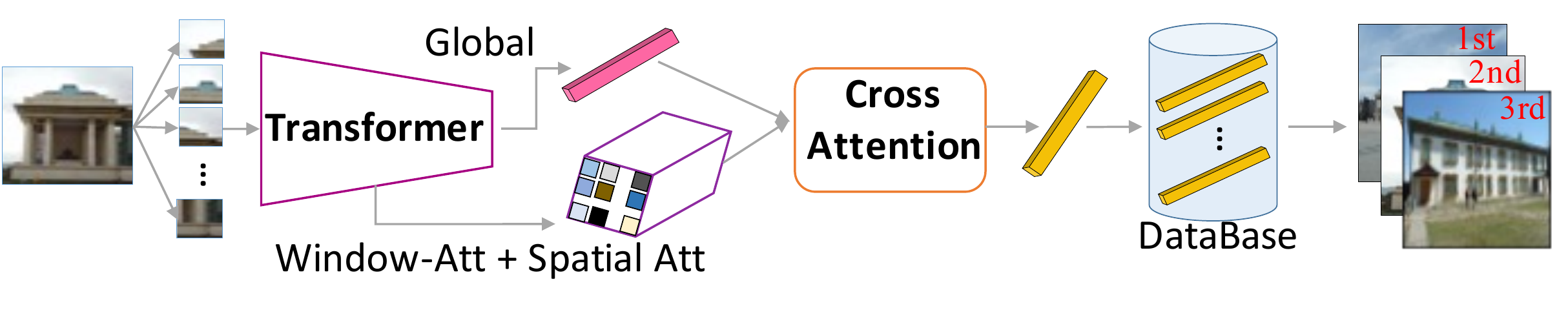} 
         \caption{Pipeline of our proposed DALG framework which works in single-scale.}
         \label{fig:dalg}
     \end{subfigure}
     \caption{Comparison between our single-scale single-stage image retrieval framework DALG and existing multi-scale two-stage or single-stage frameworks. We resort to attention for modeling global and local features and aggregating both features. With our attentive model design, efficiency can be improved via single-scale testing while maintaining the effectiveness.}
\end{figure}

Generally, we can categorize learning-based solutions into two paradigms. One stream contains various two-stage frameworks \cite{dsm,noh2017large,cao2020unifying}, among which DELG \cite{cao2020unifying} is the most representative solution as illustrated in Figure \ref{fig:delg}. Top candidates are retrieved via global descriptor and re-rank is then conducted using local feature matching. It is widely agreed that two-stage solutions are complicated and prone to suffer error accumulation. The other stream is end-to-end framework \cite{ng2020solar,arandjelovic2016netvlad,Yang_2021_ICCV}. In this paradigm, a compact descriptor is extracted from image for single-stage similarity search. Recent state-of-the-art DOLG \cite{Yang_2021_ICCV}, which explicitly blends global and local features with a manually designed orthogonal mechanism (Figure \ref{fig:dolg}), reveals that it is possible to combine global and local information for more effective retrieval without the burden of the 2nd stage. However, we can see both types of frameworks have to inference at multiple image scales to achieve robust retrieval, which is still not efficient enough. 

Naturally, we ask what is the key bottleneck that makes existing solutions have to inference at multiple image scales. 
We observe that previous solutions all base on ConvNets such as ResNet \cite{resnet}. Convolution is a local operation, therefore, ConvNets have no global access to all spatial points at each layer although stacking convolution layers can enlarge its spatial receptive field. Besides, convolution filters are applied universally for every local convolution region. As analyzed in \cite{wu2021learning}, images always show diverse local visual patterns, so universal local filtering is not capable enough to capture such local diversity.  
Currently, various Transformers variants have pushed the limits of many computer vision tasks by a large margin \cite{vit,detr,liu2021swin} owing to the excellent modeling capability of multi-head self-attention. In addition, attention has been leveraged in \cite{noh2017large,cao2020unifying,wu2021learning,Yang_2021_ICCV} to boost exploiting local image patterns. However, none of them fully explores the potential of attention.

To this end, as illustrated in Figure \ref{fig:dalg}, we propose a novel \textit{D}eep \textit{A}ttentive \textit{L}ocal and \textit{G}lobal modeling (DALG) framework for single-scale single-stage image retrieval. In detail, we leverage Swin-Transformer \cite{liu2021swin} for global feature modeling given its wonderful complexity-capability trade-off. Originated from the 2nd stage of Swin-Transformer, a local branch of multi-head window-based attention followed by spatial attention is designed. 
The window-based attention models diverse visual patterns within overlapped local regions, meanwhile the spatial attention mimics the importance sampling strategy. Finally, the hierarchical local and global features are fused with a learnable cross-attention module, rather than a handcrafted strategy of orthogonal fusion as done in DOLG \cite{Yang_2021_ICCV}. 
With our enhanced backbone design, global and local features can be well abstracted by the powerful attention mechanism, thus the learned hierarchically blending of the 2nd and the last stage features is sufficiently representative, enabling to abandon multi-scale testing. Extensive experiments on popular datasets, including Google Landmark v2 \cite{weyand2020google}, Rparis and Roxford \cite{radenovic2018revisiting}, show our solution remarkably outperforms existing solutions in terms efficiency. Meanwhile, DALG can achieve competitive accuracy.  

In a nutshell, we make the following contributions:
\begin{itemize}
    \item We concern on inefficiency of prior multi-scale testing and resort to attention mechanism 
    for efficient yet effective single-scale image retrieval.
    \item A fully attention-based DALG framework is proposed for the first time. Not only the local branch but also the global branch is based on attention to model diverse visual patterns. Besides, attention-based learnable hierarchical features fusion is introduced.
    \item Extensive experiments show DALG strikes highly promising trade-off. It remarkably improves the retrieval efficiency, meanwhile competitive mAP can be achieved compared to prior arts. 
\end{itemize}

\section{Related Work}
\label{sec:formatting}
Image retrieval has long been studied in the literature and image descriptors play the central role. Previously, handcrafted features are the main stream in this field. SIFT \cite{lowe2004distinctive} is one of the most successful scale-invariant local descriptors and SURF \cite{bay2008speeded} uses Haar filters to speed up the calculation of local features. Aggregating methods of Bow \cite{bagofwords}, VLAD \cite{vlad} and Fisher Vector \cite{sanchez2013image} are widely used to encode local features to a global descriptor for image retrieval via (approximate) nearest neighbor search. Given global retrieval results may come across unreasonable ranking among top similar candidates, a second stage re-ranking strategy with the help of RANSAC \cite{ransac} is further introduced to improve the precision \cite{avrithis2014hough,philbin2007object}.  
Later, aggregated selective match kernels (ASMK) \cite{tolias2016image} is proposed to unify aggregation-based and matching-based approaches such as Hamming Embedding \cite{jegou2008hamming}.

Deep learning has recently largely improved the progress in computer vision. Learning based image local features \cite{yi2016lift,dusmanu2019d2,detone2018superpoint,balntas2016learning,revaud2019r2d2,noh2017large,he2018local,wu2021learning} have been extensively researched. Detailed survey can be found from \cite{zhou2017recent,chen2021deep}. As a widely-known learned local feature extractor, DELF \cite{noh2017large} proposes to generate local descriptors with self-attention. In \cite{wu2021learning}, local branch is further strengthened by multiple dynamic attentions. There are also many research works proposed for global image representation learning. For example, differentiable aggregation operations of sum-pooling \cite{tolias2020learning} and GeM pooling \cite{radenovic2018fine} are invented to improve the common average pooling. Loss function optimization is also a popular research topic in terms of image representation learning, ranking-based loss functions have been promoted from triplet \cite{triplet}, quadruplet loss \cite{quadruplet} to angular loss \cite{angular} and listwise loss \cite{revaud2019learning}.  Classification based losses \cite{wang2018cosface, deng2019arcface} are also proposed to achieve more and more robust representation learning. By combining one or more of the above techniques, various network architectures \cite{babenko2014neural, babenko2015aggregating, tolias2015particular, arandjelovic2016netvlad, gordo2017end, radenovic2018fine, revaud2019learning,noh2017large,ng2020solar} are designed for discriminative image representation extraction. 
Among them, DELG \cite{cao2020unifying} unifies the local and global feature learning into one framework and largely improves the state-of-the-art performance with the two-stage retrieval then re-ranking pipeline. 

\begin{figure*}[!t]
  \centering
  \includegraphics[width=\textwidth]{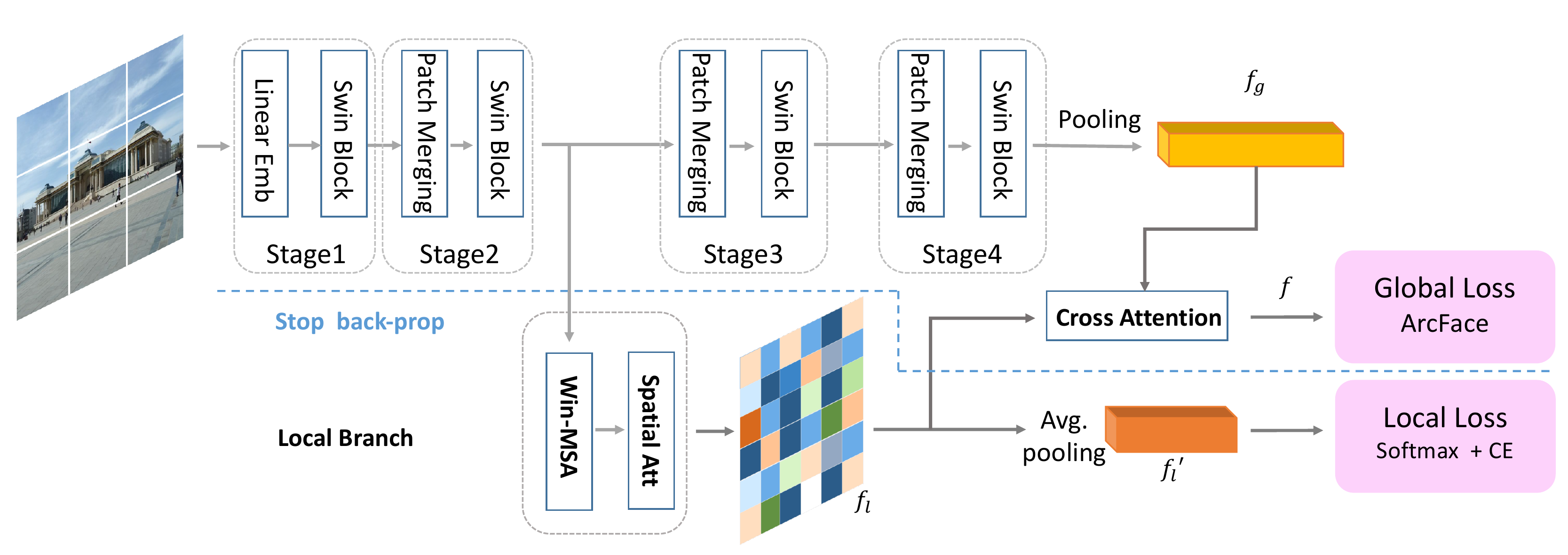} 
  \caption{Overall framework of our DALG model. Attention is fully leveraged for local and global modeling as well as fusion of these hierarchical features. Except for a global loss, an auxiliary local loss is used to help exploiting local visual patterns.}
  \label{fig:overview}
\end{figure*}
Considering that the 2nd re-ranking stage is still conducted with manually designed strategies, Transformer frameworks are recently leveraged in \cite{reranktransformer,ouyang2021contextual} to improve this procedure by learning re-ranking strategies. Although the performance of two-stage frameworks are largely improved, they are all complicated. Meanwhile, in DOLG \cite{Yang_2021_ICCV}, the authors propose to combining local and global features with a handcrafted orthogonal manner to achieve effective single-stage retrieval, thus efficiency is greatly improved. However, we find both existing two-stage and one-stage solutions take the multi-scale inference paradigm, which still limits the efficiency.  
Different from them, in this work, we attempt to fully exploit the potential of attention mechanism and focus on efficiency and effectiveness trade-off for the first time.  


\section{Methodology}

\subsection{Overall Attention-Based Framework}
The overall framework of DALG is depicted in Figure \ref{fig:overview}.
We can see that DALG follows the single-stage image retrieval paradigm proposed in DOLG \cite{Yang_2021_ICCV}, because it is an excellent start point in achieving better trade-off between efficiency and effectiveness. Our DALG is a fully attention-based design.
Specifically, we leverage Swin-Transformer \cite{liu2021swin} as our backbone instead of using ConvNets, given the following advantages which can be brought by it.
Firstly, with swin blocks, which sequentially perform window based multi-head self-attention (Win-MSA) and shifted window partition, global interactions among each local point can be efficiently modeled. Secondly, Swin-Transformer inherits similar hierarchical architecture as ResNet \cite{resnet}, hierarchical feature maps are therefore naturally available, which is of potential for us to leverage feature pyramid instead of multi-scale image pyramid. 

As Figure \ref{fig:overview} shown, global feature $f_g$ is obtained by average pooling the fourth stage of Swin-Transformer. Originated from second stage of the backbone, we devise a local branch to exploit local visual patterns. Our local branch also leverages Win-MSA considering the powerful capability of multi-head attention. Afterwards, spatial attention is used to determine importance of each local feature point and the local feature map $f_l$ is extracted. To fuse the hierarchical local and global features, cross attention module, which takes $f_g$ and $f_l$ as input, is utilized to generate the final image descriptor $f$. The cross-attention module is learnable and it can be viewed as a powerful upgrading from handcrafted orthogonal fusion strategy \cite{Yang_2021_ICCV} to learning-based fusion method driven by training data. To train our framework, we use ArcFace \cite{deng2019arcface} loss upon the final descriptor $f$. An auxiliary cross-entropy loss is applied to an temporary local feature vector $f_l'$ (which is the average pooling output of $f_l$) to help the training of the local branch. Besides, the losses of the two branches are detached such that their gradients will not back-propagate to each other. At inference, only $f$ is used for similarity matching.

\subsection{Local Branch}
\begin{figure}[t]
  \centering
  \includegraphics[width=0.8\columnwidth]{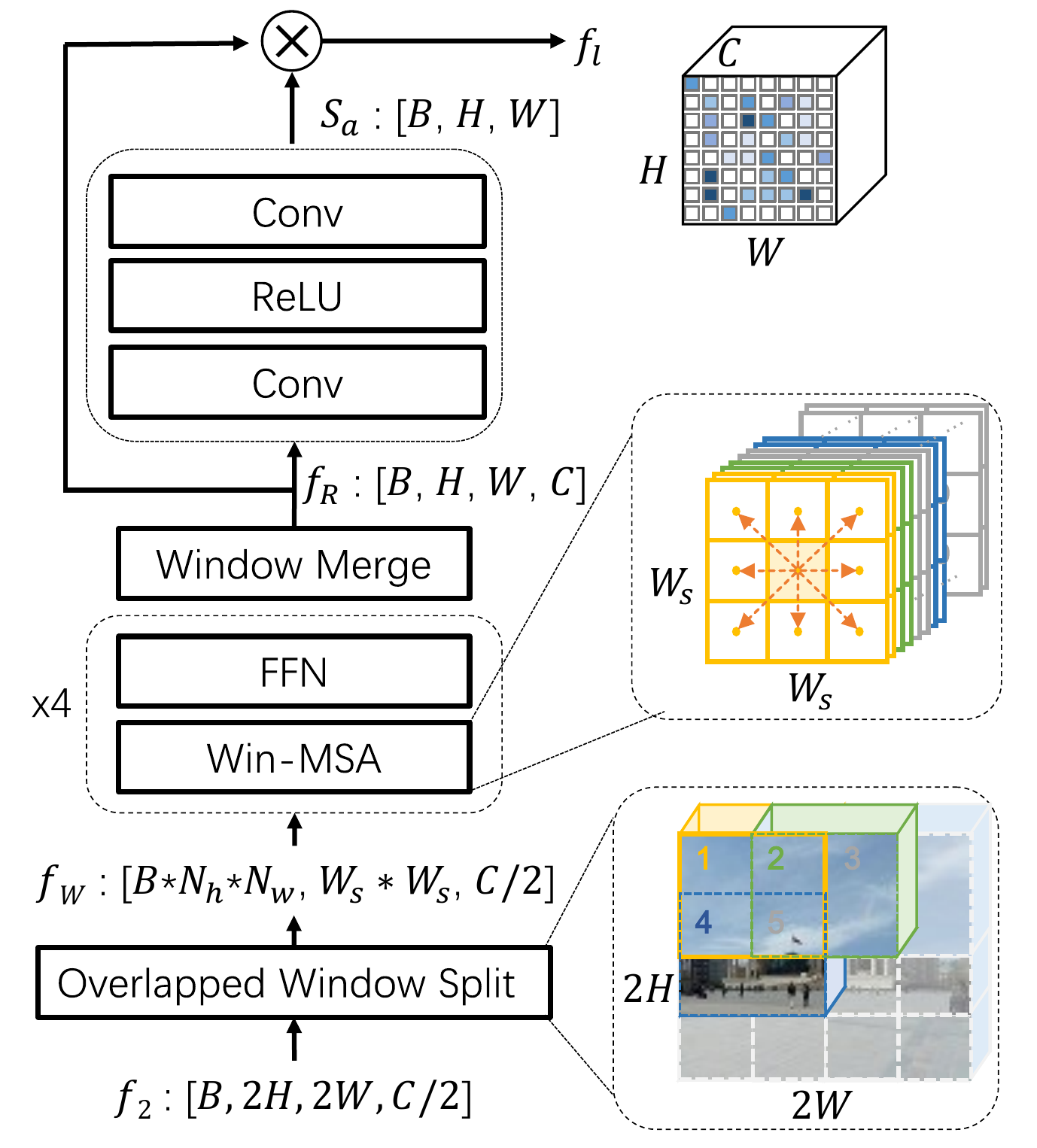} 
  \caption{Block diagram of the local branch in DALG.}
  \label{fig:local}
\end{figure}

We illustrate the network architecture of local branch in Figure \ref{fig:local}. 
Feature map $f_2\in \mathbb{R}^{B\times 2H\times 2W \times C/2}$, which is the 2nd stage output of Swin-Transformer with spatial dimension of $2H\times2W$ and channel number of $C/2$, is split into $N_h\times N_w$ local windows as $f_w$. Spatial window size is denoted as $W_s\times W_s$.  
To preserve spatial contextual structure as much as possible, we design overlapped window partition strategy, \ie, window stride is set to $s_w<W_s$. Then 4 window-based attention modules are sequentially applied to $f_w$ to exploit local visual patterns. Each window-based attention module consists of a window multi-head self-attention (Win-MSA) layer and a feed-forward network (FFN). The Win-MSA layer is responsible for conducting multi-head self-attention on each local feature window to model window-level relation, meanwhile, FFN further blends the multi-head self-attention features on per-point basis. 

Afterwards, local window features are merged at overlapped regions, reshaped to reduce spatial dimensions and then mapped to increase channel dimension, thus shaping a $B\times H\times W\times C$ feature map $f_R$ via ``Window Merge'' (similar to Patch Merge in Swin-Transformer). Then spatial attention is utilized to mimic the importance weighting process.  
In detail, its input is firstly processed using a 1 $\times$ 1 Conv-ReLU module, then the subsequent output is fed into a 1 $\times$ 1 Conv layer with SoftPlus activation to produce a spatial attention map $s_a\in \mathbb{R}^{B\times H\times W}$. Finally, the local feature $f_l$ is obtained by modulating $f_R$ with the attention map $s_a$:
\begin{equation}
 f_l = f_R*s_a   
\end{equation}

\subsection{Global and Local Fusion}
We follow the efficient yet effective single-stage art DOLG \cite{Yang_2021_ICCV}, but we propose to leverage a cross attention module to fuse the global representation $f_g$ and the local representation $f_l$. The cross attention module is learnable and we intuitively think learned fusion strategy can outperform handcrafted orthogonal fusion strategy. This is also empirically verified by our experiment. The module is build by stacking $M$ cross-attention function $\mathcal{T}: (\mathbb{R}^{1\times C}$, $\mathbb{R}^{HW\times C}) \rightarrow \mathbb{R}^{1\times C}$ to progressively select complementary information from $f_l$ to enhance $f_g$. Formally, we reshape $f_l$ to be $f_l$ $\in \mathbb{R} ^{HW\times C}$, and the final descriptor $f$ can be obtained by
\begin{equation}
\begin{split}
    &f = f_g^{(M)}, \\
    &f_g^{(m)} = \mathcal{T}_{m}(f_g^{(m-1)}, f_l), ~m\in\{1, ...,M\},\\
    &f_g^{(0)} = f_g.
\end{split}
\end{equation}

\begin{figure}[h]
  \centering
  \includegraphics[width=0.7\columnwidth]{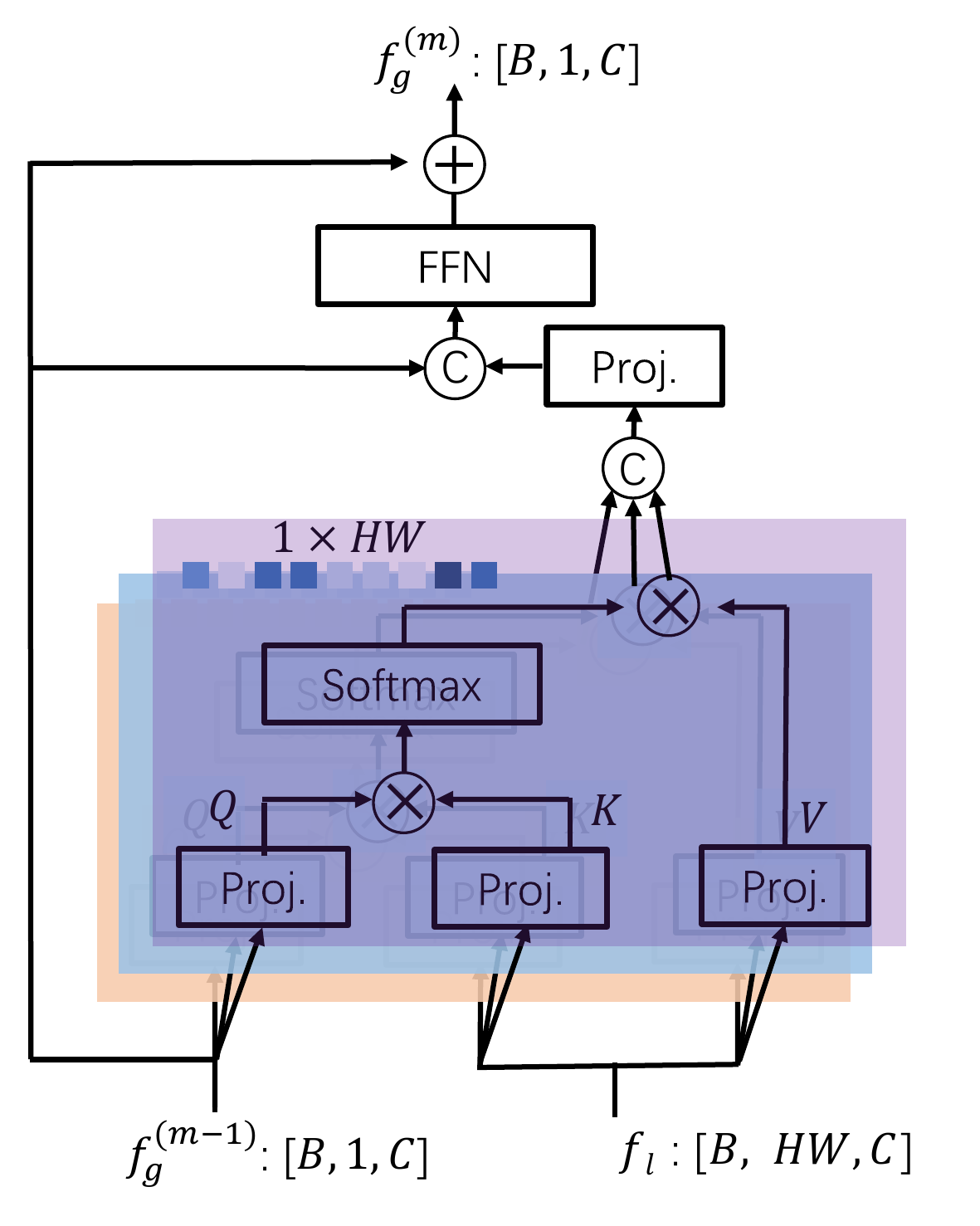} 
  \caption{Block diagram of our cross-attention function $\mathcal{T}_m$. ``Proj'' is short for projection and \textcopyright  
  ~denotes concatenation.}
  \label{fig:cross}
\end{figure}
The cross attention function $\mathcal{T}_m$ is illustrated in Figure \ref{fig:cross} and can be formulated as:
\begin{equation}
\begin{split}
    &\mathcal{T}_{m}(f_g^{(m-1)},f_l) = \\ &FFN_{m}(MCA_{m}(f_g^{(m-1)},f_l) \parallel f_g^{(m-1)})+f_g^{(m-1)},
\end{split}
\end{equation}
where $FFN_m(\cdot)$ is implemented with a two-layer feed-forward network, $MCA_m(\cdot)$ aggregates information from $f_l$ to $f_g^{(m-1)}$ with multi-head cross attention and $\parallel$ denotes concatenation. In detail, $f_g^{(m-1)}$ and $f_l$ are fused by $MCA_m(f_g^{(m-1)}, f_l)=(CA_m^{(1)}\parallel,...,\parallel CA_m^{(n_h)})W^{(m)}$, where $CA_m^{(i)}$ is the $i^{th}$ cross-attention output, $n_h$ is the number of cross-attention heads and $W^{(m)}$ is a $\mathbb{R}^{C\times C}$ projection matrix. For simplicity, let us omit the superscript $(i)$ in $CA_m^{(i)}$ in the following and $CA_m$ can be obtained by: 

\begin{equation}
\begin{split}
    Q^{(m)} &= f_g^{(m-1)}W_Q^{(m)}, \\
    K^{(m)} &= f_lW_K^{(m)}, \\
    V^{(m)} &= f_lW_V^{(m)}, \\
    CA_{m} &= softmax(\frac{Q^{(m)}{K^{(m)}}^T}{\sqrt{C/n_h}})V^{(m)}
\end{split}
\end{equation}
where $W_Q^{(m)}, W_K^{(m)}, W_V^{(m)} \in \mathbb{R}^{C\times C/n_h}$ is query, key and value projection matrix, respectively.

\section{Experiments}

\subsection{Datasets and Metrics}
\textbf{Google Landmark.}
Our models are trained on the Google Landmark v2 clean (GLDv2-clean) dataset, which is a cleaned version of the Google Landmark v2 dataset \cite{weyand2020google}. The training set of GLDv2-clean contains 1,580,470 images of 81,313 landmarks. Testing set of GLDv2-clean contains 761,757 index images and 117,577 test images.   
To evaluate our models, We follow common practice \cite{reranktransformer} and report landmark retrieval and landmark recognition results on its standard testing set. 
In detail, there are 750 test images used for private testing set. We produce retrieval predictions by ranking the index set for each of the 750 test images and evaluate the landmark retrieval performance using the mAP@100 (which indicates mean average precision evaluated at the top 100 retrieval results) metric. As for the recognition task, the official private testing set contains 76,627 images and their predictions are produced by searching from the training set. The evaluation metric of landmark recognition task is global average precision (GAP). Details of the evaluation metrics and scripts are available online\footnote{https://github.com/cvdfoundation/google-landmark}.  

We mainly train our models on google landmarks v2 clean(GLDv2-clean) dataset which is a split of GLDv2 dataset. The GLDv2-clean dataset consists of 1,580,470 images from 81,313 landmarks. For testing, we evaluate on the standard test set of GLDv2 for retrieval and recognition task. The test set contains 117577 test images and 761757 index images. In the retrieval task, 750 images are used for private solution and 379 images are used for public solution and others are ignored among the test images. Retrieval system produces predictions by searching index set with test set while recognition system produces predictions by searching training set with test set. Retrieval is evaluated using mean Average Precision@100(mAP@100), which is a variant of the standard mAP metric that only considers the top-100 ranked images. Recognition is evaluated using Global Average Precision(GAP) with one prediction per query. It is calculated by sorting all predictions in descending order of their confidence.

\textbf{ROxf and RPar.}
Revisited Oxford and Paris \cite{radenovic2018revisiting}, denoted as Roxf and Rpar in this paper, provide improved annotations for the Oxford and Paris building datasets and introduce 3 new protocols of various difficulties. Roxf and Rpar comprise 4,993 and 6,322 images respectively and they both contain 70 query images for evaluation. 
We finally report mean average precision (mAP) using the medium and hard evaluation protocols to evaluate the retrieval performance.
R1M is a large set of distractors, which consists of 1,001,001 high-resolution distracting images. The mAP results with the presence of ``+1M'' distractors is in Section. 6.

\subsection{Implementation Details}
We implement DALG variants using Swin-Transformer\cite{liu2021swin} tiny (Swin-T) and small (Swin-S) versions as backbones. Weights pre-trained on ImageNet-1K are used to initialize the backbone model parameters. As for the local branch and the cross-attention fusion module, weights are randomly initialized.  
AdamW optimizer and cosine decay of learning rate scheduler are employed to train our model. In total, 130 epochs are trained within which the first 5 epochs are trained for linear warm-up. Our models are trained with the initial learning rate of 0.00075 and 0.05 weight decay. We also retain most of the augmentation and regularization strategies of Swin-Transformer in the training phase, except Mixup, Cutmix and label smoothing. 
For the training of global branch, we use ArcFace margin loss with margin ($m$) of 0.25 and scale ($\gamma$) of 30. For local branch, softmax cross-entropy loss is used. For effectiveness, the global loss and local loss are empirically detached at $f_l$ and $f_2$, respectively. In testing phase, different from previous works \cite{cao2020unifying,Yang_2021_ICCV}, only single scale image is used for feature extraction.


The key building blocks of DALG is local branch and cross-attention fusion module. Specifically, in local branch, $f_2$ is divided into 7$\times$7 overlapped windows, the overlapping ratio is determined by the window stride $s_w$, which equals half of window size, namely $s_w=W_s/2$. The channel dimension of $f_c$ and $f_l$ is 384 and 768 respectively, that means $C=768$. The window-based multi-head attention layer in local branch contains 6 attention heads, the subsequent FFN is implemented to be two FC layers with GELU activation. The hidden size of the first FC layers is 1536 and the hidden size of the second FC layer is 384. The ``Window Merge'' module firstly combines the feature maps of the 7$\times$7 overlapped windows and reshapes them to be a $B\times 2H\times 2W \times C/2$ tensor, the features of the overlapped regions are averaged for combination. Afterwards, a spatial-to-depth operation is followed to produce a $B\times H\times W \times 2C$ tensor. Finally, in ``Window Merge'', a $1\times1$ convolution layer is applied to reduce the channel number to be $C$ and produce $f_R\in R^{B\times H\times W \times C}$. In the spatial attention part, Conv-ReLU-Conv layers are used to predict the spatial attention map, kernel size of these convolution layers is set to $1\times1$ and the second convolution layer is activated using SoftPlus, output channel number of the two convolution layers are 768 and 1, respectively. 

In the cross-attention module, there are 2 cross-attention functions stacked (\ie $M=2$) and each of them contains 12 attention heads. For each attention head, $Q, K$ and $V$ are projected via $768\times 64$ matrices. The multi-head attention outputs are concatenated and projected to be a 768 dimensional vector. The ``FFN'' in $\mathcal{T}_m$ is a two-layer perceptron and their hidden size is 1536 and 768, respectively.

Different from previous works \cite{cao2020unifying,Yang_2021_ICCV}, only single scale of image is used for feature extraction. It reduces the latency of infer process a lot which makes our DALG method an efficient and practical solution to this field especially when the dataset is large.

\subsection{Results}
\subsubsection{Comparison with State-of-the-art Methods}

\noindent\textbf{Comparison on Google Landmark.} 
\begin{table}[t]
\centering
\begin{tabular}{ll p{14mm}<{\centering}p{10mm}<{\centering}p{10mm}<{\centering}}
\toprule
\multicolumn{2}{c}{ \multirow{2}*{Method} }&{GLD version}&{Retrieval mAP\%}&{Recog. GAP\%} \\
\toprule
\multicolumn{2}{l}{R50-DELG global\cite{cao2020unifying}}&v1 &20.4 &32.4  \\ 
\multicolumn{2}{l}{R50-DELG global\cite{cao2020unifying}}&v2-clean &24.2 &27.9  \\ 
\multicolumn{2}{l}{R50-DELG\cite{cao2020unifying}}&v1 &22.3 &59.2  \\ 
\multicolumn{2}{l}{R50-DELG\cite{cao2020unifying}}&v2-clean &24.3 &55.2  \\ 
\multicolumn{2}{l}{R50-DELG+RRT\cite{reranktransformer}}&v1 &23.1 & -  \\ 
\multicolumn{2}{l}{R50-DELG+RRT\cite{reranktransformer}}&v2-clean &27.0 & -  \\ 
\toprule
\multicolumn{2}{l}{R101-ArcFace\cite{weyand2020google}}&v1 &20.6 & 33.2  \\ 
\multicolumn{2}{l}{R101-ArcFace\cite{weyand2020google}}&v2-clean &24.1 & 27.3  \\ 
\toprule
\multicolumn{2}{l}{R101-DELG global\cite{cao2020unifying}}&v1 &21.7 & 32.0  \\ 
\multicolumn{2}{l}{R101-DELG global\cite{cao2020unifying}}&v2-clean &26.0 & 28.9  \\ 
\multicolumn{2}{l}{R101-DELG\cite{cao2020unifying}}&v1 &24.3 & 61.2  \\ 
\multicolumn{2}{l}{R101-DELG\cite{cao2020unifying}}&v2-clean &26.8 & 56.4  \\ 
\toprule
\multicolumn{2}{l}{Swin-T-DALG }&v2-clean &\textbf{{30.0}} & \textbf{{62.1}} \\
\multicolumn{2}{l}{Swin-S-DALG }&v2-clean &\textbf{\underline{33.0}} & \textbf{\underline{66.5}}  \\ 
\toprule
\end{tabular}
\caption{Results (\% mAP / \% GAP) of different solutions are obtained following the retrieval and recognition evaluation protocols. All models are trained with 512 $\times$ 512 resolution. Our results are shown in the bottom and underlined results are the best.}
\label{t-1}
\end{table}

We first make a fair comparison of existing methods evaluated on the GLDv2 testing set. It will take a lot of time to evaluate on GLDv2 recognition task due to the standard protocol is to search train set with test set. However, the efficiency of our method which is single scale based end to end pipeline helps us evaluate on the recognition task quickly. Both landmark retrieval and recognition results are presented in Table \ref{t-1}. Obviously, even under the condition of single-scale testing, DALG outperforms all the other methods by a large margin. Specifically, Swin-T-DALG outperforms the recent transformer based re-ranking solution RRT \cite{reranktransformer}, which uses local descriptors from DELG \cite{cao2020unifying} for re-ranking, by 3\% in the landmark retrieval task. Compared to R101-DELG, the gain is 3.2\% in mAP and 0.9\% in GAP for retrieval and recognition tasks, respectively. 
We can also conclude from Table \ref{t-1} that landmark recognition performance is consistently better when models are trained on GLDv1, compared to training on GLDv2-clean. 
However, we observe that even if trained on GLDv2-clean, our DALG models significantly outperforms existing state-of-the-art solutions in terms of landmark recognition performance, no matter they are trained on GLDv1 or not. 
When backbone is replaced by Swin-S, DALG achieves state-of-the-art performance of 33\% in mAP for landmark retrieval and 66.5\% in GAP for landmark recognition.

\textbf{Comparison on Roxf and Rpar Datasets.} 

\begin{table*}[!htbp]
\centering
\begin{tabular}{llccccccccccc}
\toprule
\multicolumn{2}{c}{ \multirow{2}*{Method} }&\multicolumn{2}{c}{\multirow{2}*{Year}} &\multicolumn{2}{c}{\multirow{2}*{Train}}& {\multirow{1}*{Latency}} &\multicolumn{2}{c}{ Mem(GB,``+1M'')} &\multicolumn{2}{c}{Medium} &\multicolumn{2}{c}{Hard} \\
\multicolumn{6}{c}{} &(ms) &Roxf &Rpar &Roxf  &Rpar  &Roxf  &Rpar  \\

\toprule
\multicolumn{12}{l}{\textsl{(A) Local feature aggregation} } \\ \hdashline
\multicolumn{2}{l}{DELF-ASMK$\star$\cite{noh2017large, radenovic2018revisiting}} & \multicolumn{2}{c}{ICCV17} & \multicolumn{2}{c}{Landmarks} & 201 & 477.9 & 478.5 & 67.80 & 76.90 & 43.10 & 55.40 \\ 	  			 		    
\multicolumn{2}{l}{DELF-R-ASMK$\star$\cite{teichmann2019detecttoretrieve}}& \multicolumn{2}{c}{ICCV17} & \multicolumn{2}{c}{Landmarks}  & 2260 & 27.6 & 27.6 &76.00 &80.20 &52.40 &58.60 \\ 	
\multicolumn{2}{l}{{HOW-ASMK\cite{tolias2020learning}}} & \multicolumn{2}{c}{ECCV20} & \multicolumn{2}{c}{SfM-120k} & 127 & 14.3 & 14.3 &{79.40} &{81.60} &{56.90} &{62.40} \\
\multicolumn{2}{l}{{MDA-ASMK\cite{Wu_2021_ICCV}}} & \multicolumn{2}{c}{ICCV21} & \multicolumn{2}{c}{SfM-120k} & 140 & 7.4 & 7.4 &{79.50} &{81.10} &{59.20} &{62.70} \\ 
\multicolumn{2}{l}{R101-CSA\cite{Ouyang2021csa}} & \multicolumn{2}{c}{NeurIPS21} & \multicolumn{2}{c}{SfM-120k} & 212 & 7.7 & 7.7 &78.20 &88.20 &59.10 &75.30 \\ 

\toprule
\multicolumn{12}{l}{\textsl{(B) Global features} } \\ \hdashline
\multicolumn{2}{l}{{R50-DELG}\cite{cao2020unifying}} & \multicolumn{2}{c}{ECCV20} & \multicolumn{2}{c}{GLDv2-clean} & 100$^r$ & 7.6 & 7.6 &{73.60} &{81.60} &{51.00} &{71.50} \\ 
\multicolumn{2}{l}{{R101-DELG}\cite{cao2020unifying}} & \multicolumn{2}{c}{ECCV20} & \multicolumn{2}{c}{GLDv2-clean} & 145$^r$ & 7.6 & 7.6 &{76.30} &{86.60} &{55.60} &{72.40} \\ 

\toprule
\multicolumn{12}{l}{\textsl{(C) Global features + Local feature re-ranking} } \\ 

\multicolumn{2}{l}{R50-DELG\cite{cao2020unifying}} & \multicolumn{2}{c}{ECCV20} & \multicolumn{2}{c}{GLDv2-clean} & 211$^f$ & 485.5 & 486.2  &78.30 &85.70 &57.90 &71.00 \\ 
\multicolumn{2}{l}{{R101-DELG}\cite{cao2020unifying}} & \multicolumn{2}{c}{ECCV20} & \multicolumn{2}{c}{GLDv2-clean} & 383$^f$ & 485.9 & 486.6 &{81.20} &{87.20} &{{64.00}} &{72.80} \\ 
\multicolumn{2}{l}{{DELG+RRT}\cite{reranktransformer}} & \multicolumn{2}{c}{ICCV21} & \multicolumn{2}{c}{GLDv2-clean} & 571 & 243.0 & 243.3  &{78.10} &{86.70} &{60.20} &{75.10} \\ 

\toprule
\multicolumn{12}{l}{\textsl{(C) Global and Local feature fusion} } \\ \hdashline
\multicolumn{2}{l}{R50-DOLG\cite{Yang_2021_ICCV}} & \multicolumn{2}{c}{ICCV21} & \multicolumn{2}{c}{GLDv2-clean} &160 & 1.9 & 1.9 &80.50 &89.81 &58.82 &77.70 \\ 
\multicolumn{2}{l}{R101-DOLG\cite{Yang_2021_ICCV}} & \multicolumn{2}{c}{ICCV21} & \multicolumn{2}{c}{GLDv2-clean} &224 & 1.9 & 1.9 &81.50 &91.02 &61.10 &80.30 \\ 

\toprule\toprule
\multicolumn{2}{l}{Swin-T-DALG} & \multicolumn{2}{c}{ours} & \multicolumn{2}{c}{GLDv2-clean}  & \textbf{33} & 2.9 & 2.9 &\textbf{78.72} &\textbf{88.23} &\textbf{54.73} &\textbf{76.03} \\ 
\multicolumn{2}{l}{Swin-S-DALG} & \multicolumn{2}{c}{ours} & \multicolumn{2}{c}{GLDv2-clean}  & \textbf{40} & 2.9 & 2.9 & \textbf{79.94} & \textbf{90.04} & \textbf{57.55} & \textbf{79.06} \\ 
\toprule

\end{tabular}
\caption{Results (\% mAP) of different models trained with 512$\times$512 resolution. Our results are summarized in the bottom. Latency of DALG is tested on a V100 GPU and $^r$ denotes that the value is measured with our reproduction. Latency values of other methods are officially provided and tested on a P100 GPU, which is comparable with V100. ~ $^f$ denotes only latency of the global and local feature extraction of DELG is reported, latency of re-ranking by RANSAC for 1k iterations is excluded. Our DALG model is under 512$\times$512 single-scale testing. Memory is measure when inference with the presence of ``+1M'' distractors. Codes and models of DOLG \cite{Yang_2021_ICCV} are provided by its authors. }
\label{table1}
\end{table*}

\textbf{Efficiency and effectiveness trade-off.} We first provide results on Roxf and Rpar in Table \ref{table1} to compare our model with existing solutions in terms of their capabilities of balancing performance and inference complexity. For fair comparison, all models are trained with 512$\times$512 image resolution. From this table, we can see that most of previous solutions suffer from heavy memory consumption and inference latency. For example, local feature based method DELF+ASMK \cite{noh2017large} or two-stage solutions DELG \cite{cao2020unifying} and RRT \cite{reranktransformer} consume very huge memory because local descriptors have to be stored for retrieval. CSA \cite{Ouyang2021csa} can largely save memory by dynamically aggregating local descriptors. The memory usage of our DALG and DOLG \cite{Yang_2021_ICCV} are both very limited (2.9G vs. 1.9G), this is because DALG generates 768-dim descriptor rather than 512. However, efficiency of DOLG is greatly limited because it uses 5 image scales for inference. Swin-S-DALG is 4$\times$ faster than R50-DOLG, meanwhile, their mAPs are quite close (79.94 vs. 80.5). The experimental results in Table \ref{table1} evidently show the well trade-off achieved by DALG.


\vspace{-0.1cm}
\subsubsection{Comparing DALG with Other Backbones}

\begin{table}[h]
\centering
\begin{tabular}{llp{10mm}<{\centering}p{10mm}<{\centering}p{10mm}<{\centering}p{10mm}<{\centering}}
\toprule
\multicolumn{2}{c}{ \multirow{2}*{Method} }& \multicolumn{2}{c}{Roxf} &\multicolumn{2}{c}{Rpar}\\
\cline{3-4}  \cline{5-6} 
\multicolumn{2}{c}{} &M &H &M &H  \\
\toprule
	
\multicolumn{2}{l}{ViT-S\cite{vit}} &74.53 & 50.30 & 87.05 & 73.22  \\ 
\multicolumn{2}{l}{Swin-S} &76.35 & 52.73 & 87.21 & 73.95 \\ 
\toprule
\multicolumn{2}{l}{R50} &72.83 & 44.43 &84.54 &{68.81}  \\
\multicolumn{2}{l}{R50-DALG} &{73.34} &{45.28} &{84.72} &68.30\\
\multicolumn{2}{l}{Swin-T-DALG} & \textbf{76.36} & \textbf{50.81} & \textbf{86.46} & \textbf{72.67} \\ 
\multicolumn{2}{l}{Swin-S-DALG} & \textbf{\underline{78.01}} & \textbf{\underline{54.40}} & \textbf{\underline{88.97}} & \textbf{\underline{76.35}} \\ 

\toprule
\end{tabular}
\caption{Results (\% mAP) of different models trained with 256$\times$256 resolution. All methods are trained on ``GLDv2-clean'' dataset. Our final results are summarized in the bottom and the underlined numbers are the best performances.  All models in the table are under the single-scale testing paradigm.}
\label{t-backbone}
\end{table}

We also compare our model with other backbones to show the effectiveness of our proposed deep attentive local and global modeling strategy. Image resolution of $256\times256$ is used for saving training cost. First of all, we notice that many Transformer-based frameworks can naturally conduct global attentive modeling, because the multi-head attention mechanism in Transformer can well capture the feature relationship among different input patches. Therefore, Transformer can be regarded to be an off-the-shelf framework, which performs local modeling via patch-wise feed forwarding operations and global modeling via cross-patch multi-head attention. We choose ViT-S \cite{vit} and Swin-S \cite{liu2021swin} as reference solutions and train them using ArcFace loss. Results are in the top of Table \ref{t-backbone}. From these results, we can see that our DALG outperforms these existing Transformer based frameworks remarkably, this validates that our explicit attentive local and global modeling and fusion framework is a better design. Secondly, we would like to show how DALG can be used in a plug-and-play fashion for conventional convolutional network. We replace the backbone to be ResNet-50 and the local branch of DALG is appended to the \textit{res3} of ResNet-50. Then, the cross-attention module is used to fuse the global feature of ResNet-50 and the local feature produced by our local branch. The results are shown in the middle of Table \ref{t-backbone}. We can see that the proposed DALG mechanism is also applicable for convolution network. 

\vspace{-0.1cm}
\subsubsection{Ablation Study}

In this section, we conduct ablation experiments based on the Swin-S backbone. Image resolution of $256\times256$ is used for saving training cost. Results (\% mAP) are reported on Roxf and Rpar under single-scale testing.

\textbf{Local branch.} It contains overlapped Win-MSA modules and spatial attention module. We investigate how each component in the local branch will affect the performance by removing it from our model, and the experimental results are summarized in Table \ref{t-4-local-ablation}. The ``w/o Local Branch'', ``w/o Win-MSA'' and ``w/o Spatial'' refer to removing the local branch, window-based attention module and spatial attention module from DALG, respectively. The performances of these three models are inferior to the full model (DALG). And the result of 'w/o Win-MSA' shows that naively fusing the local features from $f_2$ and the global feature $f_g$ by cross attention leads to the worst performance, even worse than ``w/o Local Branch'' which means only $f_g$ is used. This is because $f_2$ is quite shallow feature, directly combining it to $f_g$ can harm the overall representation. Spatial attention can further boost the performance, for example, it gains from 77.36 to 78.01 on Roxf-M and 88.19 to 88.97 on Rpar-M. 

\begin{table}[t]
\centering
\begin{tabular}{llp{8mm}<{\centering}p{8mm}<{\centering}p{8mm}<{\centering}p{8mm}<{\centering}}
\toprule
\multicolumn{2}{c}{ \multirow{2}*{Config} }& \multicolumn{2}{c}{Roxf} &\multicolumn{2}{c}{Rpar}\\
\cline{3-4}  \cline{5-6} 
\multicolumn{2}{c}{} &M & H &M &H  \\
\toprule
\multicolumn{2}{l}{w/o Loacl Branch}& 76.36 & 49.80 & 87.78 & 74.62 \\ 
\multicolumn{2}{l}{w/o Win-MSA}&75.86 &50.36 &86.61 &73.36 \\ 
\multicolumn{2}{l}{w/o Spatial}&77.36 &53.75 &88.19 &75.72 \\ 
\multicolumn{2}{l}{Full Model (DALG)}&\textbf{78.01} &\textbf{54.40} &\textbf{88.97} &\textbf{76.35} \\ 
\toprule
\end{tabular}
\caption{ Ablation study on the local branch in our framework.}
\label{t-4-local-ablation}
\vspace{-0.1cm}
\end{table}

\textbf{Verification of the overlapped Win-MSA.}
In DALG, we use window based mulit-head self-attention with overlapped window partition to model local features. To show the importance of our design choices, we provide a comparison between different types of attention mechanisms used for exploiting local visual patterns, including multi-head self-attention (MSA) within the whole local feature map, non-overlapped window-based self-attention (N-Win-MSA) and our overlapped window-based self-attention (O-Win-MSA). 
The results can be found in Table \ref{t-4-att-ablation}. {\color{black}{From the results of first two rows, we can see that window-based multi-head attention is comparable with applying multi-head attention on the whole feature map. This is intuitively reasonable, because the local branch targets at extracting discriminative local information and window-based attention strategy can be manually forced to focus on local regions for features extraction. 
Nevertheless, attention on the whole feature map will be more homogeneous as the global branch and is more expensive, although global receptive field can be achieved.}} Actually, we cannot ensure the partition boundary will not contain critical local information, therefore overlapped partition can well alleviate the damage resulted from separated local context. With overlapped window partition, we can see O-Win-MSA outperforms N-Win-MSA by 0.53 and 0.83 on Roxf-M and Rpar-M, 0.75 and 0.51 on Roxf-H and Rpar-H, respectively.
These results also show that overlapped partition is better.

\begin{table}[t]
\centering
\begin{tabular}{llp{10mm}<{\centering}p{10mm}<{\centering}p{10mm}<{\centering}p{10mm}<{\centering}}
\toprule
\multicolumn{2}{c}{ \multirow{2}*{Method} }& \multicolumn{2}{c}{Roxf} &\multicolumn{2}{c}{Rpar}\\
\cline{3-4}  \cline{5-6} 
\multicolumn{2}{c}{} &M & H &M &H  \\
\toprule
\multicolumn{2}{l}{MSA}&77.62 &54.19 &87.92 &75.89 \\ 
\multicolumn{2}{l}{N-Win-MSA}&77.48 &53.65 &88.14 & 75.84 \\ 
\multicolumn{2}{l}{O-Win-MSA}&\textbf{78.01} &\textbf{54.40} &\textbf{88.97} &\textbf{76.35} \\ 
\hline
\end{tabular}
\caption{Experimental results of different attention modules. ``MSA'' means multi-head self-attention within the whole feature map, ``N-Win-MSA'' means MSA with non-overlapped window partition and ``O-Win-MSA'' denotes overlapped Win-MSA.}
\label{t-4-att-ablation}
\end{table}

\begin{table}[t]
\centering
\begin{tabular}{llp{8mm}<{\centering}p{10mm}<{\centering}p{10mm}<{\centering}p{10mm}<{\centering}}
\toprule
\multicolumn{2}{c}{ \multirow{2}*{Method} }& \multicolumn{2}{c}{Roxf} &\multicolumn{2}{c}{Rpar}\\
\cline{3-4}  \cline{5-6} 
\multicolumn{2}{c}{} &M & H &M &H  \\
\toprule
\multicolumn{2}{l}{ADD}&76.50 &50.24 &87.16 &73.47 \\ 
\multicolumn{2}{l}{Orthogonal\cite{Yang_2021_ICCV}}&76.88 &52.18 &88.85 &\textbf{77.31}\\
\multicolumn{2}{l}{Cross-Attention}&\textbf{78.01} &\textbf{54.40} &\textbf{88.97} &76.35 \\
\hline
\end{tabular}
\caption{Ablation study of different feature fusion methods.}
\label{t-4-orth-ablation}
\end{table}

\textbf{Fusion strategy.}
In Table \ref{t-4-orth-ablation}, we study different feature aggregation methods, including naively adding $f_l'$ to $f_g$ (ADD), recently proposed handcrafted orthogonal fusion in \cite{Yang_2021_ICCV} and the Cross-Attention fusion method proposed in this paper. From these results, we can see the proposed cross-attention fusion mechanism overall outperforms the recent state-of-the-art orthogonal fusion. With our fusion method, the mAP is improved from 76.88 to 78.01 and 88.85 to 88.97 on Roxf-M and Rpar-M, respectively. In orthogonal fusion, no parameters present and the whole strategy is handcrafted, meanwhile, cross-attention fusion can be learned from training data. The main cause of its superiority comes from the learnable multi-head cross attention functions, because data-driven solution is always more powerful than handcrafted strategies. 


\begin{table}[t]
\centering
\begin{tabular}{llp{8mm}<{\centering}p{8mm}<{\centering}p{8mm}<{\centering}p{8mm}<{\centering}}
\toprule
\multicolumn{2}{c}{ \multirow{2}*{} }& \multicolumn{2}{c}{Roxf} &\multicolumn{2}{c}{Rpar}\\
\cline{3-4}  \cline{5-6} 
\multicolumn{2}{c}{} &M & H &M &H  \\
\toprule
\multicolumn{2}{l}{W/o Stop Back-Prop}&76.35 &52.18 &86.23 &72.62 \\
\multicolumn{2}{l}{Stop CE at $f_2$}&76.47 &52.43 &87.39 &75.90 \\
\multicolumn{2}{l}{Stop ArcFace at $f_l$}& 77.46 & 53.62 & 87.12 & 75.89 \\
\multicolumn{2}{l}{Stop Both (Ours)}&\textbf{78.01} &\textbf{54.40} &\textbf{88.97} &\textbf{76.35} \\
\hline
\end{tabular}
\caption{Comparison of different training strategies. ``W/o Stop Back-prop'' means the Arcface loss of global branch and the auxiliary CE loss are both totally back-propagated. ``Stop CE at $f_2$'' means we prevent the gradients of auxiliary cross-entropy loss from back-propagating into the 2nd stage of Swin-Transformer. ``Stop ArcFace at $f_l$'' denotes the global ArcFace loss stops back-propagation at $f_l$. ``Stop Both'' is our design choice.}
\label{t-4-detach-ablation}
\end{table}

\textbf{Stop gradients.} To encourage independent local and global feature exploiting, we propose to stop the gradients of the global loss from back-propagating to the local branch and the gradients of the local loss is also detached such that it will not back-propagate to the backbone. We provide empirical results in Table \ref{t-4-detach-ablation} to validate our design choice. From this table, it can be seen that stop gradients of both local and global loss at $f_2$ and $f_l$ respectively can achieve the best mAP results. If one of the two or both stopping points are disabled, the retrieval performance degrades obviously. This empirically supports our intuition that letting the global and local branches be optimized separately can enable sufficient global and local visual patterns exploiting, and it also shows the effectiveness of our training strategy.


\section{Conclusion}
In this paper, we make the first attempt to well balance complexity and efficacy in image retrieval via a fully attention-based solution. With our model DALG, single-stage image retrieval can be further simplified to the paradigm of inference with single-scale input image. Our work successfully abandons the previous common practice of multi-scale testing, therefore, complexity can be remarkably decreased. Meanwhile, We show our method can achieve very competitive retrieval performance, owing to the proposed deep attention-based global and local modeling as well as the learnable local and global fusion strategy. Hopefully, our work will inspire the community to pay more attention on trade-off between efficiency and effectiveness.

\balance
{\small
\bibliographystyle{ieee_fullname}
\bibliography{paperbib}
}

\newpage

\appendix

\begin{figure*}[!t]
  \centering
     \begin{subfigure}[b]{\linewidth}
         \centering
         \includegraphics[width=0.95\columnwidth]{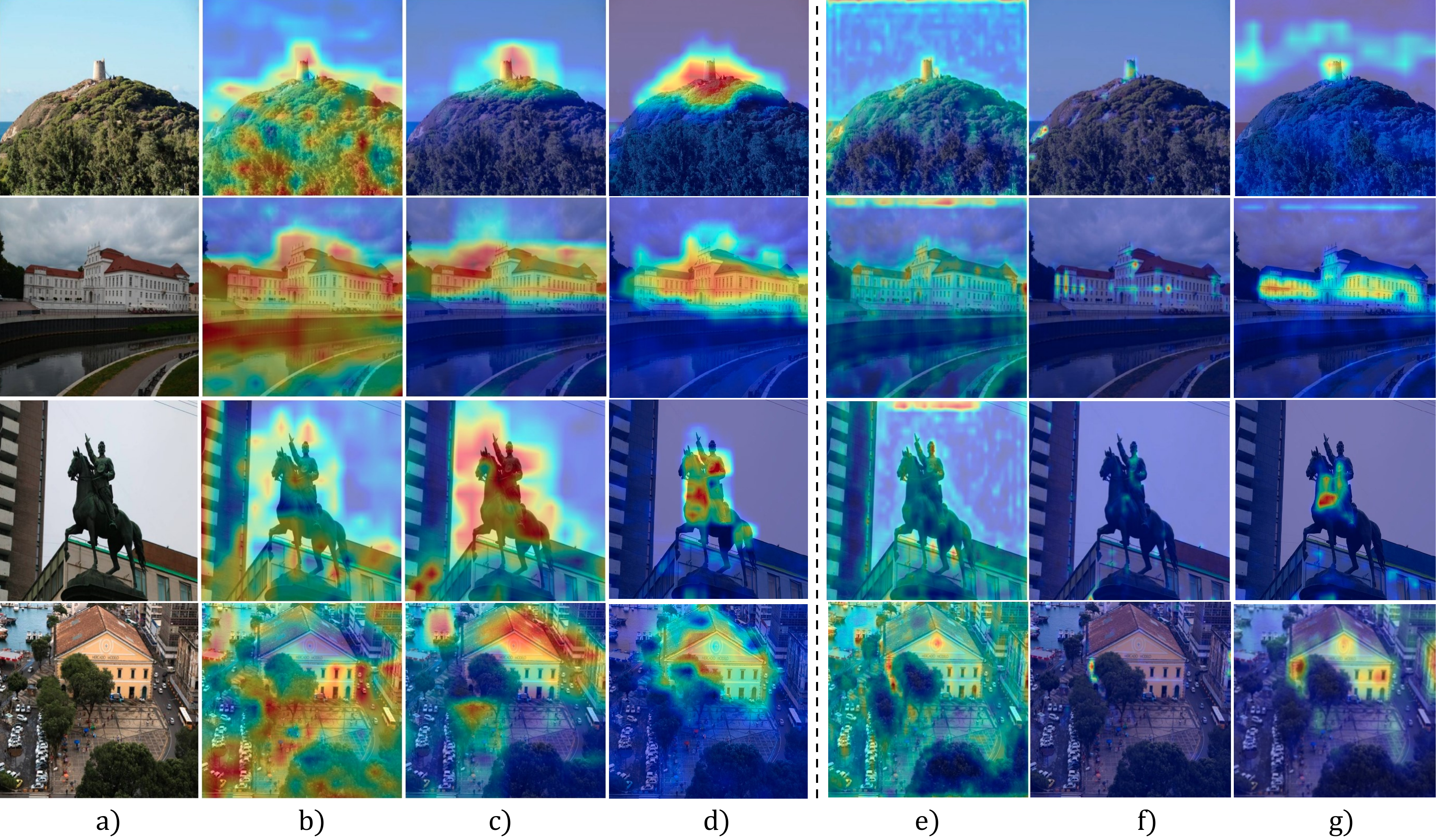} 
         \caption{}
         \vspace{0.3cm}
     \end{subfigure}
     \begin{subfigure}[b]{\linewidth}
         \centering
         \includegraphics[width=0.95\columnwidth]{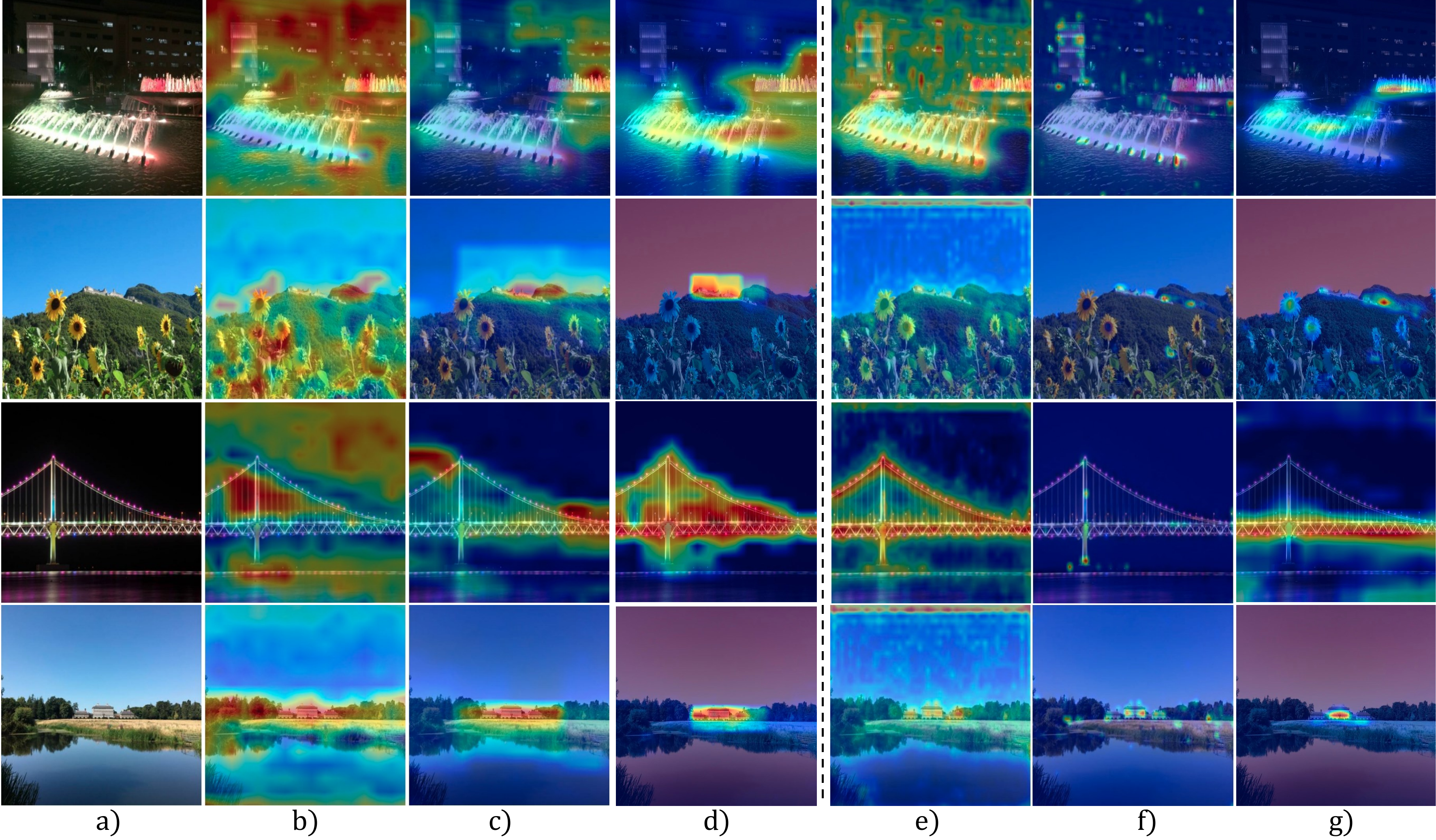} 
         \caption{}
     \end{subfigure}
     \caption{Visualization of feature activation. From left to right, a) input image; b) activation map of $f_2$, c) activation map of $f_l$ and d) activation map of $f$ in our DALG; e)activation map of $f_3$, f) activation map of local feature $f_l$, g) activation map of final descriptor $f$ in DOLG \cite{Yang_2021_ICCV}.}
     \label{fig:feat}
\end{figure*}

\begin{figure*}[!t]
  \centering
     \begin{subfigure}[b]{\linewidth}
         \centering
         \includegraphics[width=0.95\columnwidth]{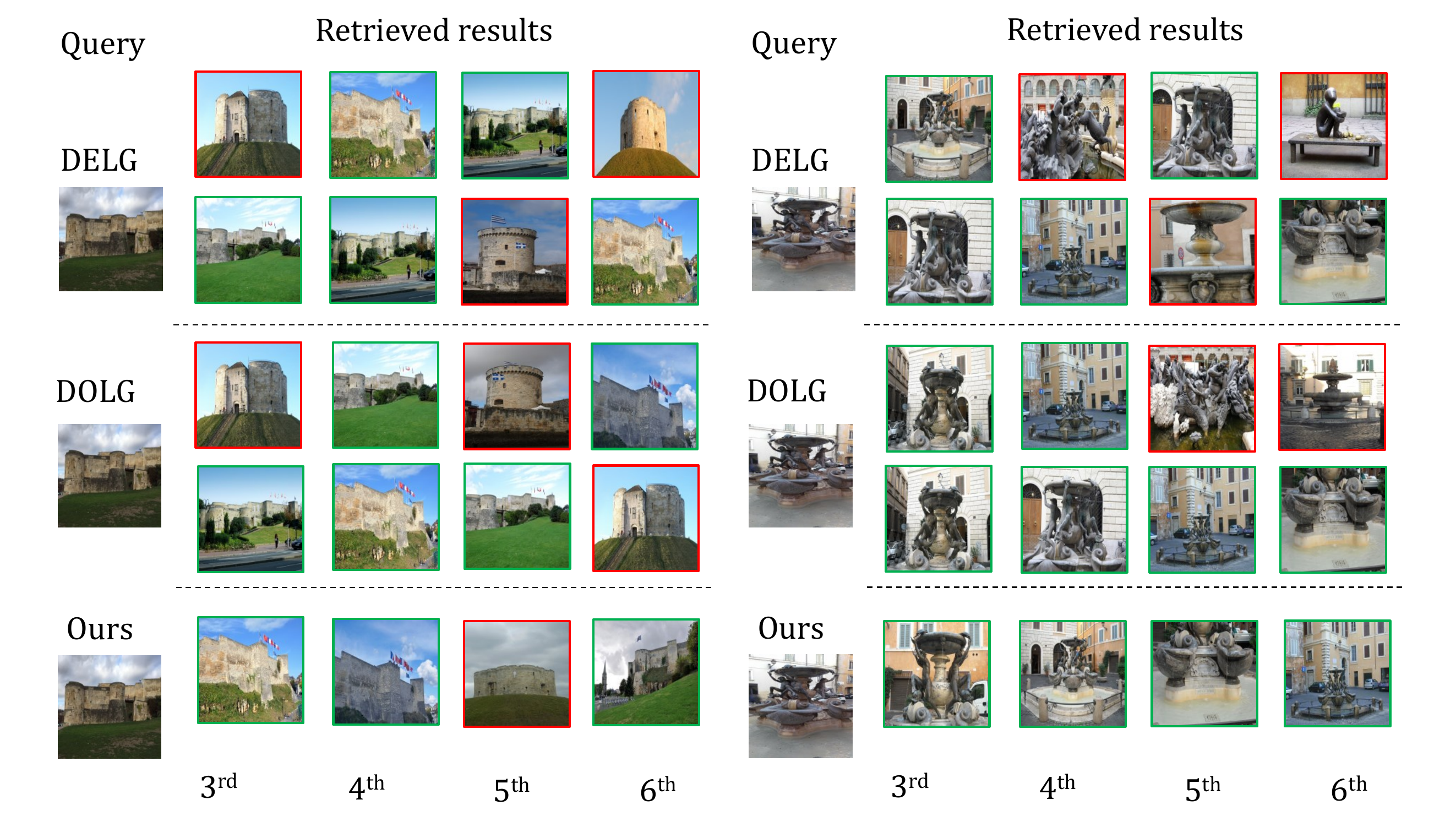} 
         \caption{}
         \vspace{0.3cm}
     \end{subfigure}
     \begin{subfigure}[b]{\linewidth}
         \centering
         \includegraphics[width=0.95\columnwidth]{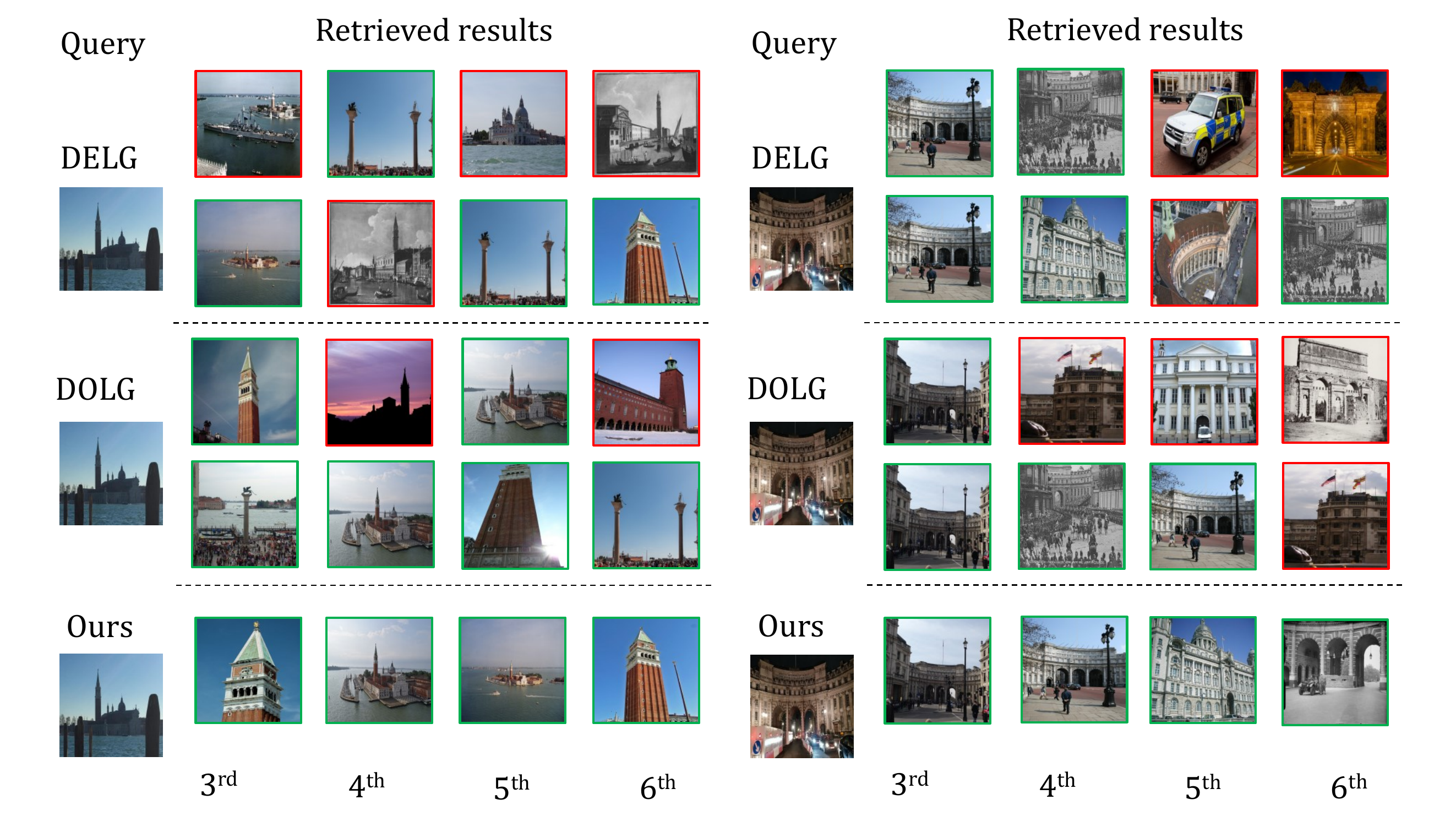} 
         \caption{}
     \end{subfigure}
     \caption{More Visualization results of the retrieval performance. We show top 3 to top 6 results of DELG$^{ss}$, DELG, DOLG$^{ss}$, DOLG, and DALG from the top to the bottom in this figure, where $^{ss}$ means the results are obtained with single-scale inference. The models are trained under image resolution of $512\times512$. Red and green boxes denote wrong and correct results, respectively.}
     \label{fig:case-show}
\end{figure*}

\begin{table}[t]
\centering
\begin{tabular}{llp{8mm}<{\centering}p{8mm}<{\centering}p{8mm}<{\centering}p{8mm}<{\centering}}
\toprule

\multicolumn{2}{c}{ \multirow{3}*{Method} }& \multicolumn{4}{c}{+1M distractor }\\
\cline{3-4}  \cline{5-6} 
\multicolumn{2}{c}{} &\multicolumn{2}{c}{Medium} &\multicolumn{2}{c}{Hard}  \\


\multicolumn{2}{c}{} &Roxf &Rpar &Roxf &Rpar  \\

\toprule
\multicolumn{6}{l}{\textsl{(A) Local feature aggregation} } \\ \hdashline
\multicolumn{2}{l}{DELF-ASMK$\star$\cite{noh2017large, radenovic2018revisiting}}  &53.80 &57.30 &31.20 &26.40\\ 	  			 		    
\multicolumn{2}{l}{DELF-R-ASMK$\star$\cite{teichmann2019detecttoretrieve}} &64.00  &59.70  &38.10  &29.40 \\ 	
\multicolumn{2}{l}{{HOW-ASMK\cite{tolias2020learning}}}   &65.80  &61.80 &38.90  &33.70 \\
\multicolumn{2}{l}{{MDA-ASMK\cite{Wu_2021_ICCV}}}  &65.60  &61.30  &42.60 &35.00 \\ 	

\toprule
\multicolumn{6}{l}{\textsl{(B) Global features} } \\ \hdashline
\multicolumn{2}{l}{{R50-DELG}\cite{cao2020unifying}}   &60.60  &68.60  &32.70 &44.40\\ 
\multicolumn{2}{l}{{R101-DELG}\cite{cao2020unifying}}   &63.70  &70.60  &37.50  &46.90\\ 

\toprule
\multicolumn{6}{l}{\textsl{(C) Global features + Local feature re-ranking} } \\ \hdashline

\multicolumn{2}{l}{R50-DELG\cite{cao2020unifying}}   &67.20  &69.60  &43.60  &45.70 \\ 
\multicolumn{2}{l}{{R101-DELG}\cite{cao2020unifying}}   &69.10  &71.50  &47.50  &48.70\\ 
\multicolumn{2}{l}{{DELG+RRT}\cite{reranktransformer}}   &67.00  &69.80  &44.10 &49.40 \\ 
\multicolumn{2}{l}{R101-CSA\cite{Ouyang2021csa}}    &61.50 &71.60 &38.20  &51.00\\ 

\toprule
\multicolumn{6}{l}{\textsl{(C) Global and Local feature fusion} } \\ \hdashline
\multicolumn{2}{l}{R50-DOLG\cite{Yang_2021_ICCV}}   &76.58 &80.79  &52.21 &62.83\\ 
\multicolumn{2}{l}{R101-DOLG\cite{Yang_2021_ICCV}}   &77.43 &83.29 &54.81 &66.69\\

\multicolumn{2}{l}{Swin-T-DALG} & 72.02 & \textbf{79.80} & 45.00 & \textbf{61.39} \\ 
\multicolumn{2}{l}{Swin-S-DALG} & \textbf{75.79} & \textbf{79.49} & \textbf{51.74} & \textbf{62.32} \\ 
\toprule
\end{tabular}
\caption{``+1M'' Results (\% mAP) of different solutions. ``$\star$'' means feature quantization is used. We use $^{ss}$ to mark the single-scale testing paradigm for DOLG. The models are trained with $512 \times 512$ training image resolution. In the most bottom part of this table, results are evaluated with $512\times512$ input resolution, meanwhile, the rest is obtained via multi-scale testing. }
\label{t:+1m}
\end{table}

\section{Results with ``+1M'' Distractors}
To evaluate the impact of distractors, we follow common practice \cite{noh2017large,cao2020unifying,Yang_2021_ICCV} to introduce the ``+1M'' dataset, which contains about 1M distracting images, to Roxf and Rpar. With presence of distractors, the performance comparison are summarized in Table \ref{t:+1m}. From this table, we can see that our DALG is competitive to DELG \cite{cao2020unifying} and DOLG \cite{Yang_2021_ICCV}. Specificially, Swin-S-DALG single scale testing achieves 75.79\% and 51.74\% on Roxf-M and Rpar-M, respectively. The mAP performance is a little bit inferior compared to the original R101-DOLG, but our Swin-S-DALG model is much more efficient than R101-DOLG (please refer to Table 1 in our main paper, Swin-S-DALG is ~6$\times$ the speed of R101-DOLG.).

\section{Visualization of Feature Activation}
We visualize the feature activation maps for our model and the recent state-of-the-art DOLG \cite{Yang_2021_ICCV}. Several results are presented in Figure \ref{fig:feat}. The left most column shows the input images, the middle part are feature activation maps extracted from our DALG model and maps from the DOLG. We can see from column b) that the 2nd stage feature $f_2$ in DALG is activated majorly by the salient regions, but there still have many irrelevant regions being activated. Meanwhile, from column c), we observe that the local features $f_l$ are more focus on the distinguishable pixels and many of the irrelevant parts are suppressed. These visualization results show that the local branch of our DALG can well model local visual patterns and dominant features can be obtained. Column d) shows the activation maps of the final compact descriptor $f$ of DALG, it can be observed that the DALG model can well attend to the landmarks. Column e) f) and g) present the activations maps of DOLG for comparison, we obtain these activation map by single-scale testing as well. The local feature $res3$ of DOLG is also majorly activated by relevant regions e), meanwhile, its local feature $f_l$ generated by DOLG's local branch f) is much more sparsely activated by image pixels when compared to c). The activation map of the final descriptor of DOLG $f$ is shown in column g), we can see from g) and d), the final descriptors generated by DALG are activated by more relevant regions and are also being less activated by irrelevant pixels. These visualization results show that our proposal deep attentive local and global modeling as well as fusion scheme can better capture the diverse visual patterns that can be leveraged for more accurate image retrieval.

\section{More Case Studies}
More image retrieval cases are demonstrated in Figure \ref{fig:case-show}. Similar observations as those in our main paper can be obtained from these results.

\section{mP@K Performance}
In order to show the percision, we further report mP@K as supplementary evaluation. From these results, we see that the precision of our proposed DALG framework is also very competitive.

\begin{table}[h]
\centering
\begin{tabular}{llp{11mm}<{\centering}p{7mm}<{\centering}p{7mm}<{\centering}p{7mm}<{\centering}p{7mm}<{\centering}}
\toprule
\multicolumn{2}{c}{ \multirow{2}*{Method} }&  \multirow{2}*{mP@k} & \multicolumn{2}{c}{Roxf} &\multicolumn{2}{c}{Rpar}\\
\cline{4-5}  \cline{6-7} 
\multicolumn{3}{c}{} &M &H &M &H  \\
\toprule

\multicolumn{2}{l}{\multirow{2}*{Swin-T-DALG}}  &k=5 &95.71 &78.86 &99.14 &94.29  \\
\multicolumn{2}{c}{} &k=10 &91.57 &68.29 &97.71 &92.00 \\
\multicolumn{2}{l}{\multirow{2}*{Swin-S-DALG}}  &k=5 &\textbf{95.71} &79.43 &\textbf{98.86} &94.29  \\
\multicolumn{2}{c}{} &k=10 &92.00 &\textbf{72.29} &\textbf{97.71} &\textbf{92.57} \\ 

\toprule
\end{tabular}
\caption{Results (\% mP@K) of DALG trained with 512$\times$512 resolution. Single-scale testing with resolution of $512\times512$ is performed.}
\label{t-mpk}
\end{table}

\end{document}